\documentclass{bmvc2k}

\usepackage{amsmath,amssymb} 
\usepackage[utf8]{inputenc} 
\usepackage[T1]{fontenc}    
\usepackage{booktabs}       
\usepackage{amsfonts}       
\usepackage{nicefrac}       
\usepackage{microtype}      
\usepackage{color}
\usepackage{tabularx}
\usepackage{arydshln}
\usepackage{multirow}
\usepackage{array}
\usepackage{bm}
\usepackage{soul}
\usepackage[export]{adjustbox}
\usepackage{graphicx}
\usepackage{verbatim}
\usepackage{caption}
\usepackage{mwe}
\usepackage[shortcuts]{extdash}
\usepackage[toc,page]{appendix}
\usepackage[normalem]{ulem}

\newcolumntype{C}[1]{>{\centering\let\newline\\\arraybackslash\hspace{0pt}}p{#1}}

\newcolumntype{O}[1]{>{\centering\let\newline\\\arraybackslash\vspace{2pt}\hspace{0pt}\vspace{0pt}}m{#1}}

\newcommand{\ie}{\textit{i}.\textit{e}., }
\newcommand{\eg}{\textit{e}.\textit{g}., }

\newcommand{\brk}[1]{{\textcolor{purple}{\sout{#1}}}}

\title{Progressive Attention Networks for Visual Attribute Prediction}

\addauthor{Paul Hongsuck Seo}{hsseo@postech.ac.kr}{1}
\addauthor{Zhe Lin}{zlin@adobe.com}{2}
\addauthor{Scott Cohen}{scohen@adobe.com}{2}
\addauthor{Xiaohui Shen}{xshen@adobe.com}{2}
\addauthor{Bohyung Han}{bhhan@snu.ac.kr}{3}

\addinstitution{
 POSTECH\\
 South Korea
}
\addinstitution{
 Adobe Research\\
 USA
}
\addinstitution{
 Seoul National University\\
 South Korea
}

\runninghead{Seo, Lin, Cohen, Shen, Han}{Progressive Attention Networks}

\def\eg{\emph{e.g}\bmvaOneDot}

\begin{document}

\maketitle


\begin{abstract}
We propose a novel attention model that can accurately attends to target objects of various scales and shapes in images.
The model is trained to gradually suppress irrelevant regions in an input image via a progressive attentive process over multiple layers of a convolutional neural network.
The attentive process in each layer determines whether to pass or block features at certain spatial locations for use in the subsequent layers.
The proposed progressive attention mechanism works well especially when combined with hard attention.
We further employ local contexts to incorporate neighborhood features of each location and estimate a better attention probability map.
The experiments on synthetic and real datasets show that the proposed attention networks outperform traditional attention methods in visual attribute prediction tasks.
\end{abstract}

\section{Introduction}
Attentive mechanisms often play important roles in modern neural networks (NNs) especially in computer vision tasks.
Many visual attention models have been introduced in the previous literature, and they have shown that attaching an attention to NNs improves the accuracy in various tasks such as image classification~\cite{STN,multiAtt,foveal}, image generation~\cite{draw}, image caption generation~\cite{showAtt} and visual question answering~\cite{askAtt,stackAtt,noh2016training}.

There are several motivations for incorporating attentive mechanisms in NNs.
One of them is analogy to the human perceptual process.
The human visual system often pays attention to a region of interest instead of processing an entire scene.
Likewise, in a neural attention model, we can focus only on attended areas of the input image.
This is beneficial in terms of computational cost; 
the number of hidden units may be reduced since the hidden activations only need to encode the region with attention~\cite{recurrentAtt}. 

\begin{figure}[t]
\centering
\begin{subfigure}[b]{0.19\linewidth}
\centering
\includegraphics[width=0.8\textwidth]{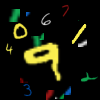}
\vspace{-0.15cm}
\subcaption{input image}
\end{subfigure}
\begin{subfigure}[b]{0.19\linewidth}
\centering
\includegraphics[width=0.8\textwidth]{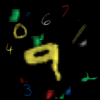}
\vspace{-0.15cm}
\subcaption{first attention}
\end{subfigure}
\begin{subfigure}[b]{0.19\linewidth}
\centering
\includegraphics[width=0.8\textwidth]{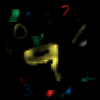}
\vspace{-0.15cm}
\subcaption{second attention}
\end{subfigure}
\vspace{0.1cm}
\begin{subfigure}[b]{0.19\linewidth}
\centering
\includegraphics[width=0.8\textwidth]{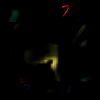}
\vspace{-0.15cm}
\subcaption{third attention}
\end{subfigure}
\begin{subfigure}[b]{0.19\linewidth}
\centering
\includegraphics[width=0.8\textwidth]{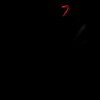}
\vspace{-0.15cm}
\subcaption{final attention}
\end{subfigure}
\vspace{-0.3cm}
\caption{
Intermediate attention maps of our progressive attention method to solve a reference problem (with query 7 and answer {\it{red}}).
It shows that attention is gradually refined through the network layers for resolving the reference problem. Distracting patterns in smaller scales are suppressed at earlier layers while those in larger scales (\eg, 9) are suppressed at later layers with larger receptive fields.
All attended images are independently rescaled for better visualization.}
\label{fig:ref_eg}
\vspace{-0.5cm}
\end{figure}

Another important motivation is that various high-level computer vision tasks should identify a particular region for accurate attribute prediction.
Figure~\ref{fig:ref_eg} illustrates an example task to predict the color (answer) of a given input number (query).
The query specifies a particular object in the input image (`7' in this example) for answering its attribute (red).
To address this type of tasks, the network architecture should incorporate an attentive mechanism either explicitly or implicitly.

One of the most popular attention mechanisms for NNs is soft attention~\cite{showAtt}, which aggregates responses in a feature map weighted by their attention probabilities.
This process results in a single attended feature vector.
Since the soft attention method is fully differentiable, the entire network can be trained end-to-end using a standard backpropagation.
However, it can only model attention to local regions with a fixed size depending on the receptive field of the layer chosen for attention.
This makes the soft attention method inappropriate for complicated cases, where objects involve significant variations in their scale and shape.


To overcome this limitation, we propose a novel attention network, referred to as \textit{progressive attention network} (PAN), which enables precise attention over objects of different scales and shapes by attaching attentive mechanisms to multiple layers within a convolutional neural network (CNN).
More specifically, the proposed network predicts attentions on intermediate feature maps and forwards the attended feature maps in each layer to the subsequent layers in CNN.
Moreover, we improve the proposed progressive attention by replacing feature aggregation, which may distort the original feature semantics, with hard attention via likelihood marginalization.
The contribution of this work is four-fold:
\begin{itemize}\setlength\itemsep{-0.05cm}
\item A novel attention model (progressive attention network) learned to handle accurate scale and shape of attentions for a target object,
\item Integration of hard attention with likelihood marginalization into the proposed progressive attention networks,
\item Use of local contexts to improve stability of the progressive attention model,
\item Significant performance improvement over traditional single-step attention models in query-specific visual attribute prediction tasks.
\end{itemize}

The rest of this paper is organized as follows. 
We first review related work in Section~\ref{relwork}. 
Section~\ref{pan} describes the proposed model with local context information. 
We then present our experimental results on several datasets in Section~\ref{exp} and conclude the paper in Section~\ref{conclusion}.

\section{Related Work}
\label{relwork}

%

\paragraph{Attention on features}
The most straightforward attention mechanism is a feature based method, which selects a subset of features by explicitly attaching an attention model to NN architectures.
This approach has improved performance in many tasks~\cite{showAtt,stackAtt,compQA,askAtt,mtatt1,mtatt2,memNet,NTM}.
For example, it has been used to handle sequences of variable lengths in neural machine translation models~\cite{mtatt1,mtatt2}, speech recognition~\cite{attASR} and handwriting generation~\cite{firstAtt}, and manage memory access mechanisms for memory networks~\cite{memNet} and neural turing machines~\cite{NTM}.
When applied to computer vision tasks to resolve reference problems, these models are designed to pay attention to CNN features corresponding to subregions in input images.
Image caption generation and visual question answering are often benefited from this attention mechanism~\cite{zheng2017learning,yu2017multi,showAtt,stackAtt,compQA,askAtt}.


\vspace{-0.3cm}
\paragraph{Attention by image transformation}
Another stream of attention model is image transformation.
This approach identifies transformation parameters fitting region of interest.
\cite{multiAtt}  and \cite{recurrentAtt} transform an input image with predicted translation parameters ($t_x$ and $t_y$) and a fixed scale factor ($\hat{s}<1$) for image classification or multiple object recognition. 
Scale factor is also predicted in \cite{draw} for image generation.
Spatial transformer networks (STNs) predict all six parameters of affine transformation, and even extend it to a projective transformation and a thin plate spline transformation~\cite{STN}. 
However, STN is limited to attending a single candidate region defined by a small number of parameters in an image.
Our model overcomes this issue by formulating attention as progressive filtering on feature maps instead of assuming that objects are roughly aligned by a constrained spatial transformation.

\vspace{-0.3cm}
\paragraph{Multiple attention processes}
There have been several approaches iteratively performing attentive processes to utilize relations between objects.
For example, \cite{stackAtt} iteratively attends to images conditioned on the previous attention states for visual question answering.
Iterative attention mechanisms to memory cells is incorporated to retrieve different values stored in the memory~\cite{memNet,NTM}.
In \cite{STN}, an extension of STN to multiple transformer layers has also been presented but is still limited to rigid shape of attention.
Our model is similar to these approaches, but aims to attend to target regions via operating on multiple CNN layers in a progressive manner; attention information is predicted progressively from feature maps through multiple layers of CNN to capture the detailed shapes of target objects.

\vspace{-0.3cm}
\paragraph{Training attention models}
The networks with soft attention are fully differentiable and thus trainable end-to-end by backpropagation.
Stochastic hard attention has been introduced in \cite{showAtt,reinforceNTM}, where networks explicitly select a single feature location based on the predicted attention probability map.
Because the explicit selection (or sampling) procedure is not differentiable, hard attention methods employ REINFORCE learning rule~\cite{reinforce} for training.
Transformation-based attention models~\cite{multiAtt,recurrentAtt} are typically learned by REINFORCE as well while STN~\cite{STN} proposes a fully differentiable formulation suitable for end-to-end learning.
The proposed network has advantage of end-to-end trainability by a standard backpropagation without any extra technique. 

\setlength{\abovedisplayskip}{1pt}
\setlength{\belowdisplayskip}{1pt}

\vspace{-0.2cm}

\section{Progressive Attention Networks (PANs)}
\vspace{-0.1cm}
\label{pan}
To alleviate the limitations of existing attention models in handling variable object scales and shapes, we propose a progressive attention mechanism. 
We describe technical details about our attention model in this section.

%
\begin{figure*}[t]
\centering
\includegraphics[width=1\linewidth] {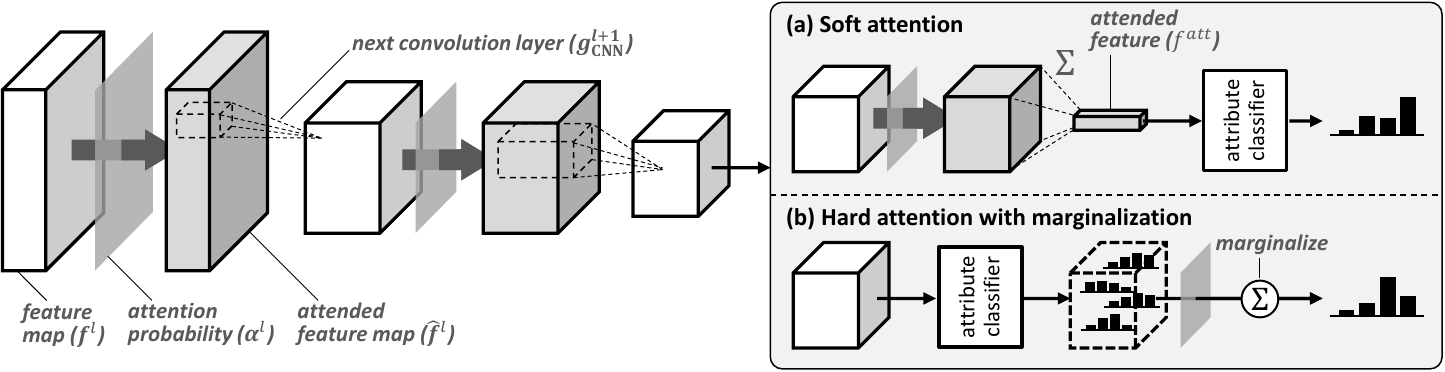}
\vspace{-0.15cm}
\caption{Overall procedure of our progressive attention model.
Attentive processes are successively applied to feature maps at multiple layers and the resulting attended feature maps are used as input feature maps for the next convolution layers in the CNN.
Attention probabilities $\alpha^l$ are estimated from feature maps and input query.
In the last attention layer, (a) the attended feature maps are aggregated to a single feature vector (by sum pooling) as in soft attention and fed to the final attribute classifier, or (b) attention probabilities are used to marginalize attribute predictions from hard-selected feature at each spatial location as in hard attention.
The bar graphs in dotted cube denote attribute prediction results from individual spatial locations and $\Sigma$ in circle represents marginalization with attention probabilities.
}
\label{fig:architecture}
\vspace{-0.3cm}
\end{figure*}

\vspace{-0.2cm}
\subsection{Progressive Attentive Process}

Let ${\bm{f}}^l \in \mathbb{R}^{H_l\times W_l\times C_l}$ be an output feature map of a layer $l \in \{0, \dots, L\}$ in CNN with width $W_l$, height $H_l$ and number of channels $C_l$, and $f^l_{i,j} \in \mathbb{R}^{C_l}$ be a feature at $(i, j)$ of feature map ${\bm{f}}^l$.
In the proposed model, an attentive process is applied to multiple layers of CNN and we obtain the following attended feature map ${\bm{\hat{f}}}^l = [ \hat{f}_{i,j}^l  ]$, where
\begin{equation}
\hat{f}^l_{i,j}=\alpha^l_{i,j} f^l_{i,j}.
\end{equation}
In the above equation, attention probability $\alpha^l_{i,j}$ for a feature $f^l_{i,j}$ and a query $q$ are given by
\begin{align}
s^l_{i, j} = g^l_\mathrm{att}(f^l_{i, j}, q; \bm{\theta}^l_\mathrm{att}), ~~~~~\text{and}~~~~~
\alpha_{i,j}^l = 
	\begin{cases}
		\mathrm{softmax}_{i,j}(\bm{s}^l) & \text{if} ~~ l=L \\
		\sigma(s_{i,j}^l) & \text{otherwise}
	\end{cases}
\label{eq:att}
\end{align}
where $g^l_\mathrm{att}(\cdot)$ denotes the attention function with a set of parameters $\bm{\theta}^l_\mathrm{att}$, $s^l_{i, j}$ is the attention score at $(i, j)$, and $\sigma(\cdot)$ is a sigmoid function.
The attention probability at each location is estimated independently in the same feature map; a sigmoid function is employed to constrain attention probabilities between 0 and 1. 
For the last layer of attention, we use a softmax function over the entire spatial region for later feature aggregation.

Unlike the soft attention model~\cite{showAtt}, the attended feature map ${\bm{\hat{f}}}^l$ in the intermediate attention layers is not summed up to generate a single vector representation of the attended regions.
Instead, the attended feature map is forwarded to the next layer as an input to compute the next feature map, which is given by
\begin{equation}
{\bm{f}}^{l+1} = g^{l+1}_\mathrm{CNN}(\bm{\hat{f}}^l; \bm{\theta}^{l+1}_\mathrm{CNN}) 
\label{eq:attended_feature_map_recursion}
\end{equation}
where $g^{l+1}_\mathrm{CNN}(\cdot)$ is the convolution operation at layer $l+1$ in CNN parameterized by $\bm{\theta}^{l+1}_\mathrm{CNN}$.
This feedforward procedure with attentive processes in CNN is performed from the input layer, where ${\bm{f}}^{0} = I$, until ${\bm{\hat{f}}}^L$ is obtained.
The attended feature $f^\mathrm{att}$ is finally retrieved by summing up all the features in the final attended feature map ${\bm{\hat{f}}}^L$ as in soft attention, which is given by
\begin{equation}
f^\mathrm{att}= \sum_i^H{\sum_j^W{\hat{f}^L_{i,j}}}= \sum_i^H \sum_j^W \alpha^L_{i, j}f^L_{i, j}.
\label{eq:att_feat}
\end{equation}
The attended feature $f^\mathrm{att}$ obtained by this process is then used as the input to visual attribute classifier as shown in Figure~\ref{fig:architecture}a.

In our models, we place the attention layers to the output of max pooling layers instead of every layer in CNN because the reduction of feature resolution within CNN mainly comes from pooling layers.
In practice, we can also skip the first few pooling layers and attach the attention module only to the outputs of last $K$ pooling layers.

\vspace{-0.3cm}
\subsection{Incorporating Hard Attention}
\vspace{-0.1cm}
The derivation of the final attended feature $f^\mathrm{att}$ shown in Eq.~\eqref{eq:att_feat} is similar to soft attention mechanism.
However, since $f^\mathrm{att}$ is obtained by a weighted sum of features at all locations, it loses semantic layout of the original feature map. 
To overcome this limitation, we extend our model to incorporate hard attention.
Given the final feature map $\bm{f}^L$, the hard attention model predicts an answer distribution using the selected feature $f^L_{i,j}$ as
\begin{equation}
    p(a|\bm{f}^L, i, j) = H(f^L_{i,j})
\end{equation}
where $a$ is the predicted answer and $H(\cdot)$ is the visual attribute classifier.
The final answer distribution given the feature map $\bm{f}^L$ is obtained by
\begin{align}
    p(a|\bm{f}^L) = \sum_{i}^{H}\sum_{j}^{W}{p(a|\bm{f}^L, i, j)p(i, j|\bm{f}^L)} = \sum_{i}^{H}\sum_{j}^{W}{H(f^L_{i,j})\alpha^L_{i,j}}
\end{align}
where $p(i, j|\bm{f}^L)$ is modeled by $\alpha^L_{i,j}$ in Eq.~\eqref{eq:att}.
This marginalization eliminates the need for REINFORCE technique used in \cite{showAtt} by removing the hard selection process, and helps maintain the unique characteristics of feature maps.

\vspace{-0.3cm}
\subsection{Multi-Resolution Attention Estimation}
\vspace{-0.1cm}
As shown in Eq.~\eqref{eq:att}, the resolution of attention probability map $\bm{\alpha}^l$ depends on the size of the feature map in the corresponding layer.
Since the resolution of $\bm{\alpha}^l$ decreases inherently with increasing depth of CNNs and the attentive processes are performed over multiple layers recursively in our framework, we can attend to the regions with arbitrary resolutions even in higher layers.
Hence, the proposed network exploits high-level semantics in deep representations for inference without losing attention resolution. 

The progressive attention model is also effective in predicting fine attention shapes as attention information is aggregated over multiple layers to suppress irrelevant structures at different levels. 
In lower layers, features whose receptive fields contain small distractors are suppressed first while the features from a part of large distractors remain intact.
In higher layers, features corresponding to these large distractors would have low attention probability as each feature contains information from larger receptive fields enabling the attention module to distinguish between distractors and target objects.
This phenomenon is demonstrated well in the qualitative results in our experiments (Section~\ref{exp}).
%
An additional benefit of our progressive attention is that inference is based only on feedforward procedure.

%
\begin{figure}[t]
\centering
\includegraphics[width=0.6\linewidth] {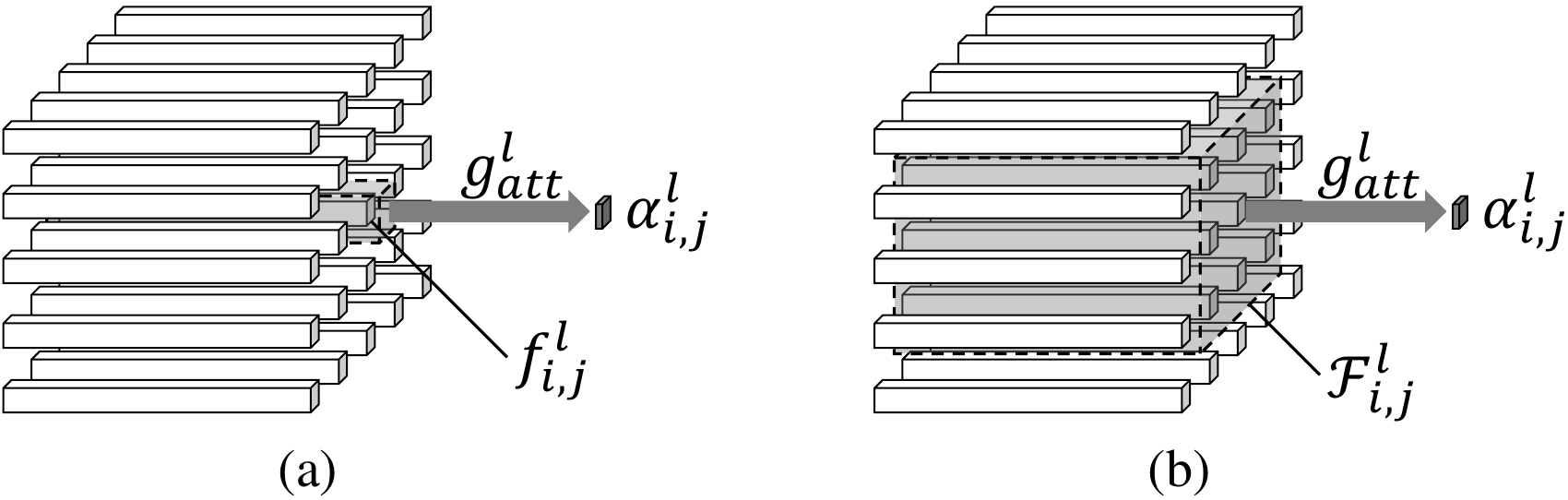}
\vspace{-0.2cm}
\caption{Attention estimation (a) without and (b) with local context. In (a), $\alpha^l_{i,j}$ is predicted from $f^l_{i,j}$ only while its spatially adjacent features are also employed to estimate $\alpha^l_{i,j}$ in (b).}
\label{fig:local_context}
\vspace{-0.5cm}
\end{figure}

\vspace{-0.3cm}
\subsection{Local Context}
\vspace{-0.1cm}
PAN can improve the quality of attention estimation by allowing its attention layers to observe a local context of the target feature.
%
The local context $\mathcal{F}^l_{i, j}$ of a feature $f^l_{i,j}$ is composed of its spatially adjacent features as illustrated in Figure~\ref{fig:local_context}, and is formally denoted by $\mathcal{F}^l_{i,j}=\{f^l_{s,t} | i-\delta \le s \le i+\delta, j-\delta \le t \le j+\delta \}$.
The attention score is now predicted by the attention network with local context as
\begin{equation}
s^l_{i, j} = g^l_{\mathrm{att}}(\mathcal{F}^l_{i, j}, q; \bm{\theta}^l_{\mathrm{att}}).
\end{equation}
In our architecture, the area of the local context is given by the filter size corresponding to the composite operation of convolution followed by pooling in the next layer.
The local context does not need to be considered in the last layer of attention since its activations are used to compute the final attended feature map.
Local context improves attention prediction quality by comparing centroid features with their surroundings and making the estimated attention more discriminative.

\vspace{-0.3cm}
\subsection{Training Progressive Attention Networks}
\vspace{-0.1cm}

Training a PAN is as simple as training a soft attention network~\cite{showAtt} because every operation within the network is differentiable.
The entire network with both soft and hard attention is trained end-to-end by the standard backpropagation minimizing the cross entropy of the object-specific visual attributes.
When we train it from a pretrained CNN, the CNN part should always be fine-tuned together since the intermediate attention maps may change the input distributions of their associated layers in the CNN.

\vspace{-0.3cm}
\section{Experiments}
\vspace{-0.2cm}
\label{exp}
This section discusses experimental results on two datasets.
Note that our experiments focus on the tasks directly related to visual attention to minimize any artifacts caused by irrelevant components.
The codes are available at \href{https://github.com/hworang77/PAN}{https://github.com/hworang77/PAN}.

\begin{figure}[t]
\centering
\begin{subfigure}[b]{0.19\linewidth}
\centering
\includegraphics[width=0.8\linewidth]{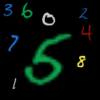}
\vspace{-0.1cm}
\subcaption{MREF}
\label{fig:MREF}
\end{subfigure}~~~~~~
\begin{subfigure}[b]{0.19\linewidth}
\centering
\includegraphics[width=0.8\linewidth]{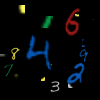}
\vspace{-0.1cm}
\subcaption{MDIST}
\label{fig:MDIST}
\end{subfigure}~~~~~~
\begin{subfigure}[b]{0.19\linewidth}
\centering
\includegraphics[width=0.8\linewidth]{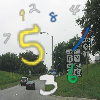}
\vspace{-0.1cm}
\subcaption{MBG}
\label{fig:MBG}
\end{subfigure}
\label{fig:ex_dataset}
\caption{Example of the MREF datasets.}
\vspace{-0.5cm}
\end{figure}

\vspace{-0.3cm}
\subsection{MNIST Reference}
\vspace{-0.1cm}
\paragraph{Datasets}
We conduct experiments on synthetic datasets created from MNIST~\cite{MNIST}.
The first synthetic dataset is referred to as MNIST Reference (MREF; Figure~\ref{fig:MREF}), where each training example is a triple of an image, a  query number, and its color label.
The task on this dataset is to predict the color of the number given by a query.
Five to nine distinct MNIST digits with different colors out of $\{\text{green},\text{yellow},\text{white},\text{red},\text{blue}\}$ and scales in $[0.5, 3.0]$ are randomly sampled and located within a $100\times100$ empty image with black background.
When coloring numbers, Gaussian noise is added to the reference color value.
To simulate more realistic situations, we made two variants of MREF by adding distractors (MDIST; Figure~\ref{fig:MDIST}) or replacing background with natural images (MBG; Figure~\ref{fig:MBG}).
Distractors in MDIST are constructed with randomly cropped $5\times5$ patches of MNIST images whereas backgrounds of MBG are filled with natural scene images randomly chosen from the SUN Database~\cite{SUN}.
The training, validation and test sets contain 30K, 10K and 10K images, respectively.

\begin{figure}[t]
\centering
\begin{subfigure}{0.5\textwidth}
\centering
\includegraphics[width=0.9\textwidth]{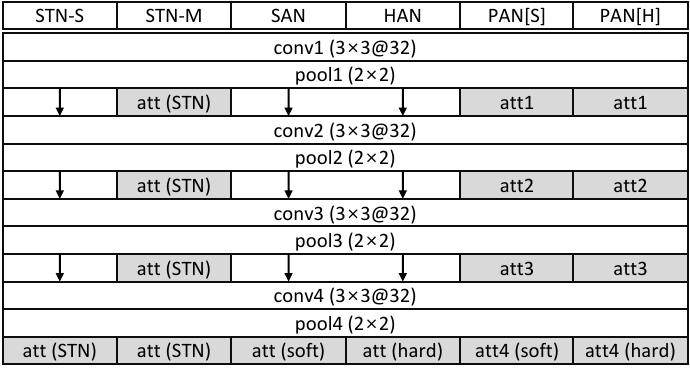}
\vspace{-0.1cm}
\subcaption{Network architectures of models on MREF. Arrows represents direct connection to next layer without attention.}
\label{fig:net_architecture}
\vspace{-0.2cm}
\end{subfigure}~~~~~~~
\begin{subfigure}{0.4\textwidth}
\centering
\includegraphics[width=0.9\textwidth]{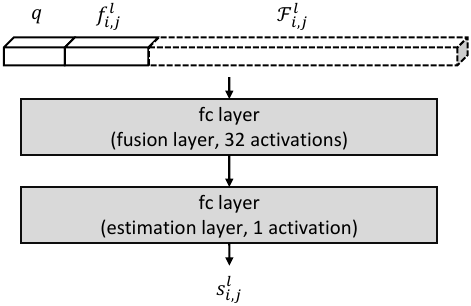}
\vspace{-0.1cm}
\subcaption{Architecture of attention function $g^l_{\mathrm{att}}(\cdot)$.  Local contexts $\mathcal{F}^l_{i, j}$ are used only in PAN[*]+CTX.}
\label{fig:att_architecture}
\end{subfigure}
\vspace{0.3cm}
\caption{
Detailed illustration of network architectures on MNIST Reference experiments. 
}
\label{fig:network_details}
\end{figure}

\paragraph{Experimental setting}

We implement two variants of the proposed network with soft and hard attention at its final layer, denoted by PAN[S] and PAN[H], respectively.
Each of them is implemented with and without the local context observation; the networks with local context observation are denoted by `+CTX' in our results. 
In addition, soft attention network (SAN), hard attention network (HAN)~\cite{showAtt}, and two variants of spatial transformer network (STN-S and STN-M)~\cite{STN} are used as baseline models for comparisons.
While STN-S is the model with a single transformer layer, STN-M contains multiple transformer layers in the network.
We reimplemented SAN and STNs following the descriptions in \cite{showAtt} and \cite{STN}, respectively, and trained HAN by optimizing the marginal log-likelihood.
The architecture of image encoding networks in SAN and HAN, and localization networks in STNs are all identical for fair comparisons.
The CNN in the proposed network also has the same architecture except for the additional layers for progressive attention.
The CNN is composed of four stacks of $3\times3$ convolutions with 32 channels (stride 1) followed by a $2\times2$ max pooling layer (stride 2) as illustrated in Figure~\ref{fig:net_architecture}.
We used a single fc layer for attribute classifier because the task requires simple color prediction. 
The attention functions $g^l_{\mathrm{att}}(\cdot)$ for all models are given by multi-layer perceptrons with two layers (Figure~\ref{fig:att_architecture}).
The function takes the concatenation of a query vector $q$ and a feature vector $f^l_{i,j}$, and outputs an attention score $s^l_{i,j}$, where the query vector $q$ is a one-hot vector representing a target object.
In the proposed models, the intermediate attention functions additionally take local context $\mathcal{F}^l_{i, j}$ containing spatially adjacent features with $\delta=1$.
Every model is trained end-to-end from scratch by RMSprop until the models show no improvement for 50 epochs.
We exponentially decay the initial learning rate by 0.9 at every 30 epoch after the 50th epoch, and grid-search the best initial learning rate for each model in [0.001, 0.005].

%

\begin{table}[t]
\centering
\caption{Results on MREF datasets: (left) color prediction accuracy [\%] and true-positive ratio [\%]. (right) accuracy [\%] with different scales on MBG test subset.}
\label{tab:MNIST_acc}
\vspace{0.3cm}
\scalebox{0.7}{
\begin{tabular}{
@{}C{2.2cm}@{}|@{}C{1.1cm}@{}@{}C{1.1cm}@{}|@{}C{1.1cm}@{}@{}C{1.1cm}@{}|@{}C{1.1cm}@{}@{}C{1.1cm}@{}
}
		    & \multicolumn{2}{c|}{MREF} & \multicolumn{2}{c|}{MDIST}    & \multicolumn{2}{c}{MBG}   \\ 
            & Acc.      & TPR		    & Acc.      & TPR		        & Acc.      & TPR		    \\ \hline
Uniform     & 20.00     & 2.34		    & 20.00     & 2.35              & 20.00     & 2.39          \\ \hline
STN-S 	    & 39.10		& 0.94		    & 38.32		& 0.52              & 32.27	    & 0.74          \\
STN-M	    & 93.89		& 0.66		    & 85.09		& 0.74              & 52.25	    & 0.65          \\ \hline
SAN		    & 82.94		& 13.83		    & 75.73		& 12.82             & 53.77	    & 6.92          \\
HAN		    & 81.84		& 13.95		    & 78.49		& 13.81             & 55.84	    & 7.64          \\
PAN[S]	    & 95.92		& 44.25		    & 91.65		& 52.85             & 69.46	    & 32.07         \\
PAN[H]	    & 95.32     & 62.31         & 90.46		& 52.07             & 72.50	    & 38.32         \\
PAN[S]+CTX  & 98.51		& 60.45         & \bf96.02	& 59.60             & 81.01	    & 43.87         \\
PAN[H]+CTX  & \bf98.53	& \bf62.36      & 95.82		& \bf61.39          & \bf84.62	& \bf51.49      \\ \hline
\end{tabular}
}
\scalebox{0.7}{
\begin{tabular}{
@{}C{2.2cm}@{}|@{}C{1.3cm}@{}@{}C{1.3cm}@{}@{}C{1.3cm}@{}@{}C{1.3cm}@{}@{}C{1.3cm}@{}
}
		    & \multicolumn{5}{c}{Scale ranges}                     \\ 
            & 0.5-1.0   & 1.0-1.5       & 1.5-2.0   & 2.0-2.5       & 2.5-3.0   \\ \hline
STN-S 	    & 31.1		& 35.3		    & 35.8		& 37.7          & 45.0	    \\
STN-M	    & 51.8		& 58.3		    & 55.3		& 58.5          & 61.7	    \\ \hline
SAN		    & 49.1		& 68.9		    & 63.6		& 61.7          & 54.6	    \\
HAN		    & 51.2		& 68.7		    & 70.8		& 64.0          & 59.5	    \\
PAN[S]	    & 67.9		& 76.6		    & 73.9		& 69.6          & 61.3	    \\
PAN[H]	    & 70.8      & 79.6          & 76.8		& 70.2          & 67.3	    \\
PAN[S]+CTX  & 79.6		& 86.8          & 85.1      & \bf81.9       & 72.1	    \\
PAN[H]+CTX  & \bf84.2	& \bf87.1       & \bf87.5	& 81.6          & \bf79.6	\\ \hline
\end{tabular}
}
\vspace{-0.1cm}
\end{table}

\begin{figure*}[t]
\centering
\includegraphics[width=1\linewidth] {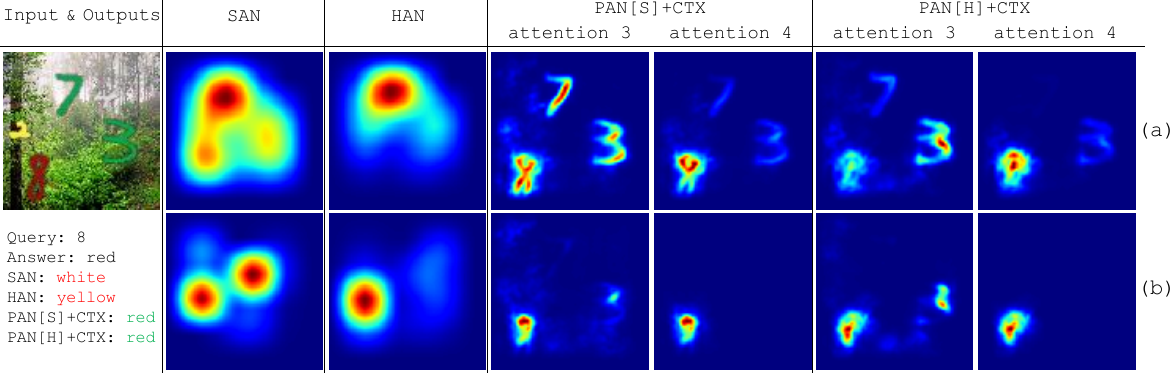}
\vspace{-0.5cm}
\caption{
Qualitative results of SAN, HAN, PAN[S]+CTX and PAN[H]+CTX. 
(a) Magnitude of activations in feature maps $f^l_{i,j}$ before attention; the activations are mapped to original image space by spreading activations to their receptive fields.
(b) Magnitude of activations in attended feature maps $\hat{f}^l_{i,j}$ showing the effect of attention in contrast to (a).
For PAN[$*$]+CTX, we only show last two attentions, which accumulate the attentions of earlier layers.
Every map is rescaled into $[0, 1]$ by $(x-\min)/(\max-\min)$.
}
\label{fig:MREF_qualitative}
\vspace{-0.3cm}
\end{figure*}

\paragraph{Results}
Table~\ref{tab:MNIST_acc}(left) presents color prediction accuracy of all compared algorithms.
PAN[$*$] clearly outperforms other approaches with significant margins and PAN[$*$]+CTX further improves performance by exploiting local context for attention estimation.
Our progressive attention model with the hard attention process and local context achieves the best performance in most cases.
%
%
%
Note that the attention capability of the baseline models is restricted to either rhombic or coarsely shaped regions.
In contrast, the proposed networks predict attention maps with arbitrary shapes by capturing spatial support of target area better.

We also present the attention quality of the models using true-positive ratio (TPR) in Table~\ref{tab:MNIST_acc}(left).
TPR measures how strong attention is given to proper location by computing the ratio of the aggregated attention probability within the desired area (\ie ground-truth segmentation) to the attention probability in the whole image.
To calculate TPR of STN baselines, we assigned the uniform attention probabilities to the attended rhombic regions.
The models with progressive attention give the best results with significant margins compared to all the other methods.
These results suggest that progressive attention constructs more accurate shapes of attended regions than all other attention models.
Integrating hard attention and local context further improves overall performance in most cases.

To evaluate scale sensitivity, we divide test sets into five subsets based on target object scales with uniform intervals and computed accuracies of the models.
In Table~\ref{tab:MNIST_acc}(right), the results on MBG show that the models with progressive attention are robust to scale variations due to their multi-scale attention mechanism especially when the hard attention and local contexts are incorporated.
The tendencies on MREF and MDIST are also similar.


An important observation regarding to STNs is that these models actually do not attend to target objects.
Instead, STNs generate almost identical transformation regardless of input images and pass the filtering process to next layers.
As the results, the transformed images are padded ones containing the entire original image with different affine transformations. 
Consequently, these models show very low TPRs, even lower than the uniform attention.

Figure~\ref{fig:MREF_qualitative} illustrates qualitative results of the two proposed methods and two baselines on MBG dataset. 
Our models yield accurate attended regions by gradually augmenting attention and suppressing irrelevant regions in the image.
We observe that the proposed models maintain high attention resolution through the progressive attention process.
In contrast, the baseline models attend to target regions only once at the top layer resulting in coarse attention. 

\subsection{Attribute prediction on Visual Genome}
\vspace{-0.1cm}
\paragraph{Dataset}
Visual Genome (VG)~\cite{VisGen} is an image dataset containing several types of annotations: question/answer pairs, image captions, objects, object attributes and object relationship. 
We formulate object attribute prediction as a multi-label classification task with reference.
Given an input image and a query (\ie an object category), we predict binary attributes of individual objects specified by the query.
We used 827 object classes and 749 attribute classes that appear more than 100 times.
A total of 86,674 images with 667,882 object attribute labels are used for our experiment, and they are split into training, validation and test sets, which contain 43,337, 8,667 and 34,670 images, respectively.
The task is challenging because appearances and semantics of objects largely vary.

\begin{table}[t]
\centering
\caption{Weighted mAP of the attribute prediction and TPR of attentions measured with ground-truth bounding boxes on VG dataset (left). TPR of attentions measured with ground-truth segmentations in VOC 2007 (right).} 
\vspace{0.2cm}
\label{tab:VisGen_result}
\scalebox{0.7}{
\begin{tabular}[m]{
@{}C{2.5cm}@{}|@{}C{1.5cm}@{}@{}C{1.5cm}@{}|@{}C{1.5cm}@{}@{}C{1.5cm}@{}
}
			& \multicolumn{2}{c|}{attention only} 			& \multicolumn{2}{c}{w/ prior}  \\ 
			& mAP 		& TPR		& mAP		& TPR		\\ \hline
STN-S		& 28.87		& 11.59		& 30.50		& 7.78		\\
STN-M		& 29.12		& 1.99		& 31.17		& 2.28		\\ \hline
SAN			& 27.62		& 15.01		& 31.84		& 17.65		\\
HAN			& 27.72		& 17.24		& 31.93		& 19.70		\\
PAN[S]+CTX	& 29.38 	& 18.01 	& 32.50 	& 20.17 	\\
PAN[H]+CTX	& \bf{30.00}& \bf{22.23}& \bf{34.19}& \bf{24.37}\\ \hline
\end{tabular}
}~~~~~
\scalebox{0.7}{
\begin{tabular}[m]{
@{}C{2.5cm}@{}|@{}C{2.8cm}@{}
}
            & TPR (VOC 2007)    \\ \hline
SAN			& 22.01		        \\
HAN         & 24.91             \\
PAN[S]+CTX  & 27.16             \\
PAN[H]+CTX  & \bf31.79          \\ \hline
\end{tabular}
}
\vspace{-0.2cm}
\end{table}
\paragraph{Experimental settings and results}
All networks share VGG-16 network~\cite{VGG16} pretrained on ImageNet~\cite{ImageNet} and is further fine-tuned for attribute prediction.
For SAN and HAN, an attention layer is attached to the last pooling layer of VGG-16 while PAN[S]+CTX and PAN[H]+CTX stack an additional attention layer with local contexts $\mathcal{F}^l_{i, j}$ with $\delta=2$ on top of the last three pooling layers in VGG-16.
We skip the first two pooling layers ($\mathrm{pool1}$ and $\mathrm{pool2}$) for placing attention because the features in those layers are not discriminative enough to filter out. 
We also test the models with object class conditional prior.
For the purpose, the final attended feature is fused with the query once more using a fully connected layer, which allows the network to reflect the conditional distribution of the attributes given the query.
We train the models by Adam with different module-wise initial learning rates and the initial learning rates are exponentially decayed by the factor of 0.9.
Refer to the supplementary document and the code for more detailed descriptions on the network architectures and experimental settings, respectively.

All models are evaluated via mean average precision (mAP) weighted by the frequencies of the attribute labels in test set, where the computation of mAP follows PASCAL VOC protocol~\cite{voc}.
Our progressive attention process with local context and hard attention mechanism consistently achieves the best weighted mAP scores in both experimental settings as shown in Table~\ref{tab:VisGen_result}(left). 
Table~\ref{tab:VisGen_result}(left) also presents TPR of each model measured with ground-truth bounding boxes due to lack of the ground-truth segmentation labels for evaluating attention qualities.
PAN[H]+CTX shows the best TPR although the computation of TPR with bounding boxes is more favorable to other methods.
%
\begin{figure*}[t]
\centering
\includegraphics[width=1\linewidth] {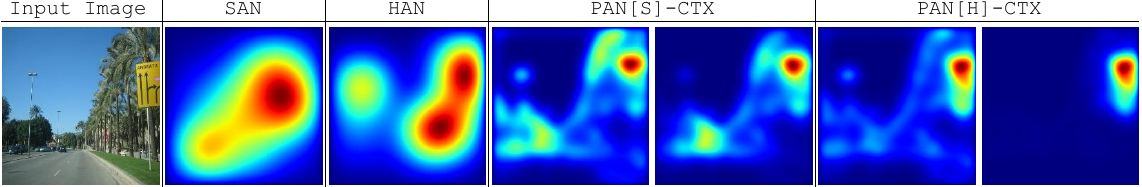}
\caption{
Visualization of example attentions of models on VG dataset.
Two variants of progressive attention models gradually attend to target objects (queried by {\it{sign}}) in fine resolution. For PAN[$*$]+CTX, we only show last two attentions, which accumulate attentions of earlier layers.
More qualitative results are presented in supplementary document.
}
\label{fig:VG_qualitative}
\vspace{-0.5cm}
\end{figure*}


To measure the attention quality with segmentation masks, we also measure TPRs of models on train/val set in PASCAL VOC 2007~\cite{voc2007} as shown in Table~\ref{tab:VisGen_result}(right).
We use class names as queries for images to obtain an attention map of each image and measure TPR with its corresponding object segmentation mask.
We observe that our progressive attention models outperform the baselines and become stronger when incoporated with hard attention.

%
STNs have trouble to attend to target objects as in MREF and show very poor TPRs.
Note that STNs show higher mAPs than the other baselines in `attention only' setting.
We believe that this is because STNs utilize object class conditional priors by encoding queries through a manipulation in the transformation process.
Figure~\ref{fig:VG_qualitative} presents qualitative results on VG dataset, which show that the progressive attention models gradually attend to target objects.


\vspace{-0.3cm}
\section{Conclusion}
\vspace{-0.1cm}
\label{conclusion}
We proposed a novel attention network, which progressively attends to regions of interest through multiple layers in a CNN.
As the model is recursively applied to multiple layers of a CNN with an inherent feature hierarchy, it  accurately predicts regions of interest with variable sizes and shapes.
We also incorporate local contexts into our attention network for more robust estimation.
While progressive attention networks can be implemented with either soft or hard attention, we demonstrated that both versions of the model substantially outperform existing methods on both synthetic and real datasets.

\section*{Acknowledgement}
This research is partly supported by Adobe Research and Institute for Information \& communications Technology Promotion (IITP) grant funded by the Korea government (MIST) (No. 2017-0-01778, Development of Explainable Human-level Deep Machine Learning Inference Framework; No.2017-0-01780, The Technology Development for Event Recognition/Relational Reasoning and Learning Knowledge based System for Video Understanding).

\bibliography{egbib}

\clearpage
\begin{appendices}

\section{Network Architectures on Visual Genome}

\begin{figure}[t]
\centering
\includegraphics[width=0.75\linewidth]{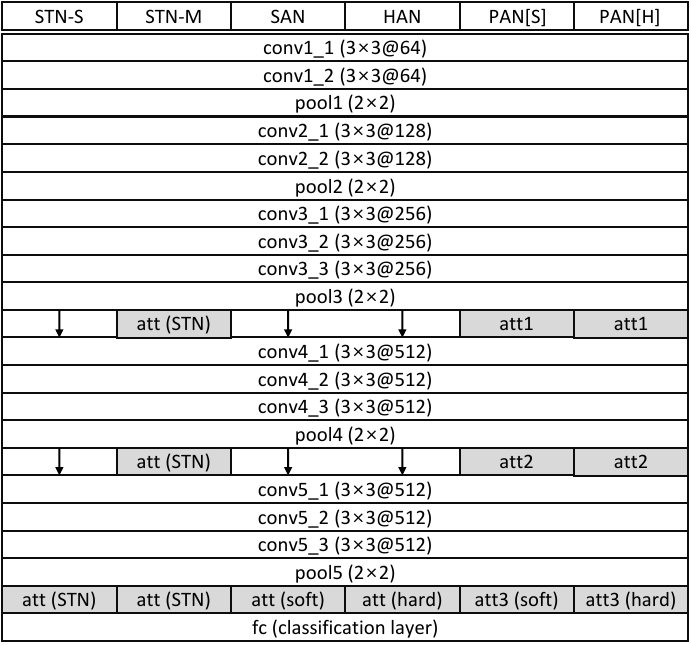}
\caption{Network Architectures of Models.}
\label{fig:vg_net_architecture}
\end{figure}

\begin{figure}[t]
\centering
\begin{subfigure}[b]{0.4\linewidth}
\includegraphics[width=1\linewidth]{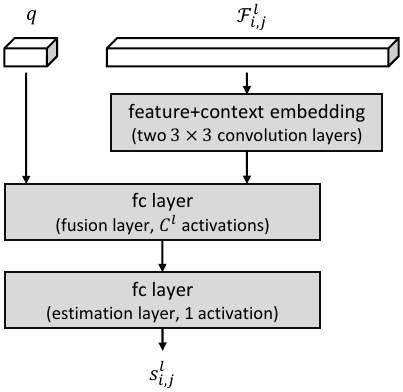}
\subcaption{}
\label{fig:vg_att_architecture1}
\end{subfigure} ~~~~~~
\begin{subfigure}[b]{0.35\linewidth}
\includegraphics[width=1\linewidth]{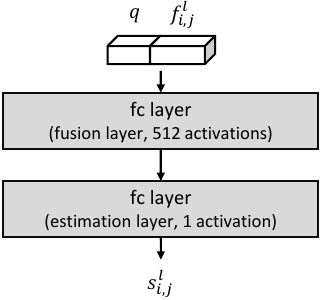}
\subcaption{}
\label{fig:vg_att_architecture2}
\end{subfigure}
\caption{(a) Architecture of the intermediate attention functions $g^l_{\mathrm{att}}(\cdot)$ in $\mathrm{att1}$ and $\mathrm{att2}$ of PAN, and (b) architecture of the attention functions of SAN and HAN, and the last attention function of PAN.}
\end{figure}

In PAN, the convolution and pooling layers of VGG-16 network~\citep{VGG16}, pretrained on ImageNet~\citep{ImageNet}, are used, and three additional attention layers $\mathrm{att1}$, $\mathrm{att2}$ and $\mathrm{att3}$ are stacked on top of the last three pooling layers $\mathrm{pool3}$, $\mathrm{pool4}$ and $\mathrm{pool5}$ respectively as illustrated in Figure~\ref{fig:vg_net_architecture}. 
The attention functions of $\mathrm{att1}$ and $\mathrm{att2}$ take the local contexts $\mathcal{F}^l_{i,j}$ in addition to the query $q$ and the target feature $f^l_{i,j}$ to obtain the attention score $s^l_{i,j}$.
The size of the local contexts is squared with that of the receptive fields of the next two convolution layers before the next attention by setting $\delta=2$.
Two convolutions same as the next two convolution layers in CNN firstly encode the target feature and the local context, and are initiallized with the same weights as in CNN (Figure~\ref{fig:vg_att_architecture1}).
This embedding is then concatenated with the one-hot query vector and fed to two fully connected layers, one fusing two modalities and the other estimating the attention score.
In $\mathrm{att3}$, the attention function takes the concatenation of the query and the target feature and feed it to two fully connected layers (Figure~\ref{fig:vg_att_architecture2}).
The attended feature $f^\mathrm{att}$ obtained from the last attention layer $\mathrm{att3}$ is finally fed to a classification layer to predict the attributes.

The baseline networks also share the same architecture of CNN of VGG-16 network as in PAN (Figure~\ref{fig:vg_net_architecture}). 
The soft attention and the hard attention is attached to the top of CNN instead in SAN and HAN, respectively.
The attention functions in the baselines consist of two fully connected layers taking the concatenation of the query and the target feature as in the attention function of  $\mathrm{att3}$ in PAN.

The proposed network and the baselines described above use the query for obtaining the attention probabilities and give us the pure strength of the attention models.
However, the target object class, represented by the query, gives much more information than just attetion.
It confines possible attributes and filters irrelevant attributes.
For these reasons, we additionally experiment on a set of models that incorporate the target object class conditional prior for the attribute prediction.
In these models, the query is fused with the attended feature $f^\mathrm{att}$ by an additional fully connected layer and the fused feature is used as the input of the classification layer.

\label{VG_architecture}

\clearpage
\section{More Qualitative Results on MBG}
\begin{table}[h!]
    \centering
    \begin{tabular}{
    @{}O{2cm}@{}|@{}O{2cm}@{}@{}O{2cm}@{}@{}O{2cm}@{}@{}O{2cm}@{}|@{}O{2.1cm}@{}
    }
        Input image & (a) Masked image & (b) Before attention & (c) After attention & (d) Original resolution & Models \\ \hline
        \includegraphics[width=0.95\linewidth,height=0.95\linewidth]{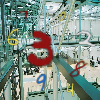} &
        \includegraphics[width=0.95\linewidth,height=0.95\linewidth]{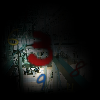} &
        \includegraphics[width=0.95\linewidth,height=0.95\linewidth]{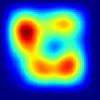} &
        \includegraphics[width=0.95\linewidth,height=0.95\linewidth]{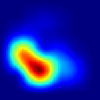} &
        \includegraphics[width=0.95\linewidth,height=0.95\linewidth]{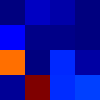} &
        SAN \\ \hline
        &
        \includegraphics[width=0.95\linewidth,height=0.95\linewidth]{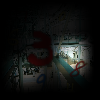} &
        \includegraphics[width=0.95\linewidth,height=0.95\linewidth]{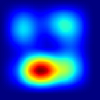} &
        \includegraphics[width=0.95\linewidth,height=0.95\linewidth]{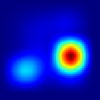} &
        \includegraphics[width=0.95\linewidth,height=0.95\linewidth]{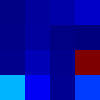} &
        HAN \\ \cline{2-6}
        &
        \includegraphics[width=0.95\linewidth,height=0.95\linewidth]{figures/supplementary/mref/2/han_img.png} &
        \includegraphics[width=0.95\linewidth,height=0.95\linewidth]{figures/supplementary/mref/2/han_bef.png} &
        \includegraphics[width=0.95\linewidth,height=0.95\linewidth]{figures/supplementary/mref/2/han_aft.png} &
        \includegraphics[width=0.95\linewidth,height=0.95\linewidth]{figures/supplementary/mref/2/han_att.png} &
        HAN \\ \cline{2-6}
        &
        \includegraphics[width=0.95\linewidth,height=0.95\linewidth]{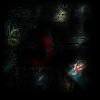} &
        \includegraphics[width=0.95\linewidth,height=0.95\linewidth]{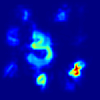} &
        \includegraphics[width=0.95\linewidth,height=0.95\linewidth]{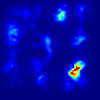} &
        \includegraphics[width=0.95\linewidth,height=0.95\linewidth]{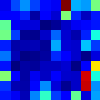} &
        PAN[S]+CTX (attention 3)\\ \cline{2-6}
        &
        \includegraphics[width=0.95\linewidth,height=0.95\linewidth]{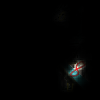} &
        \includegraphics[width=0.95\linewidth,height=0.95\linewidth]{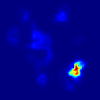} &
        \includegraphics[width=0.95\linewidth,height=0.95\linewidth]{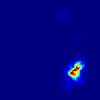} &
        \includegraphics[width=0.95\linewidth,height=0.95\linewidth]{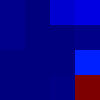} &
        PAN[S]+CTX (attention 4)\\ \cline{2-6}
        &
        \includegraphics[width=0.95\linewidth,height=0.95\linewidth]{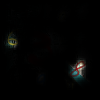} &
        \includegraphics[width=0.95\linewidth,height=0.95\linewidth]{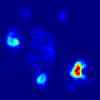} &
        \includegraphics[width=0.95\linewidth,height=0.95\linewidth]{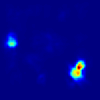} &
        \includegraphics[width=0.95\linewidth,height=0.95\linewidth]{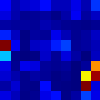} &
        PAN[H]+CTX (attention 3)\\ \cline{2-6}
        &
        \includegraphics[width=0.95\linewidth,height=0.95\linewidth]{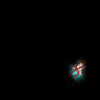} &
        \includegraphics[width=0.95\linewidth,height=0.95\linewidth]{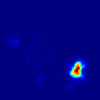} &
        \includegraphics[width=0.95\linewidth,height=0.95\linewidth]{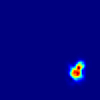} &
        \includegraphics[width=0.95\linewidth,height=0.95\linewidth]{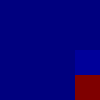} &
        PAN[H]+CTX (attention 4)\\ \cline{2-6}
    \end{tabular}
    \vspace{0.3cm}
    \caption{
Qualitative results of SAN, HAN, PAN[S]+CTX and PAN[H]+CTX with query '8'. 
(a) Input images faded by attended feature map (c).
(b) Magnitude of activations in feature maps $f^l_{i,j}$ before attention; the activations are mapped to original image space by spreading activations to their receptive fields.
(c) Magnitude of activations in attended feature maps $\hat{f}^l_{i,j}$ showing the effect of attention in contrast to (b).
For PAN[$*$]+CTX, we only show last two attentions, which accumulate the attentions of earlier layers.
Every map is rescaled into $[0, 1]$ by $(x-\min)/(\max-\min)$.
}
    \label{tab:my_label}
\end{table}

\begin{table}[h!]
    \centering
    \begin{tabular}{
    @{}O{2cm}@{}|@{}O{2cm}@{}@{}O{2cm}@{}@{}O{2cm}@{}@{}O{2cm}@{}|@{}O{2.1cm}@{}
    }
        Input image & Masked image & Before att. & After att. & Org. resolution & Models \\ \hline
        \includegraphics[width=0.95\linewidth,height=0.95\linewidth]{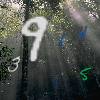} &
        \includegraphics[width=0.95\linewidth,height=0.95\linewidth]{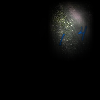} &
        \includegraphics[width=0.95\linewidth,height=0.95\linewidth]{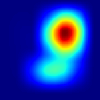} &
        \includegraphics[width=0.95\linewidth,height=0.95\linewidth]{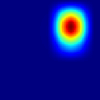} &
        \includegraphics[width=0.95\linewidth,height=0.95\linewidth]{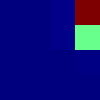} &
        SAN \\ \hline
        &
        \includegraphics[width=0.95\linewidth,height=0.95\linewidth]{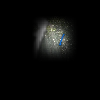} &
        \includegraphics[width=0.95\linewidth,height=0.95\linewidth]{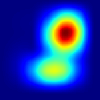} &
        \includegraphics[width=0.95\linewidth,height=0.95\linewidth]{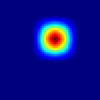} &
        \includegraphics[width=0.95\linewidth,height=0.95\linewidth]{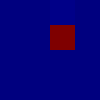} &
        HAN \\ \cline{2-6}
        &
        \includegraphics[width=0.95\linewidth,height=0.95\linewidth]{figures/supplementary/mref/1/han_img.png} &
        \includegraphics[width=0.95\linewidth,height=0.95\linewidth]{figures/supplementary/mref/1/han_bef.png} &
        \includegraphics[width=0.95\linewidth,height=0.95\linewidth]{figures/supplementary/mref/1/han_aft.png} &
        \includegraphics[width=0.95\linewidth,height=0.95\linewidth]{figures/supplementary/mref/1/han_att.png} &
        HAN \\ \cline{2-6}
        &
        \includegraphics[width=0.95\linewidth,height=0.95\linewidth]{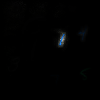} &
        \includegraphics[width=0.95\linewidth,height=0.95\linewidth]{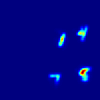} &
        \includegraphics[width=0.95\linewidth,height=0.95\linewidth]{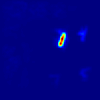} &
        \includegraphics[width=0.95\linewidth,height=0.95\linewidth]{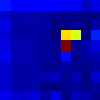} &
        PAN[S]+CTX (attention 3)\\ \cline{2-6}
        &
        \includegraphics[width=0.95\linewidth,height=0.95\linewidth]{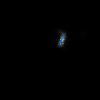} &
        \includegraphics[width=0.95\linewidth,height=0.95\linewidth]{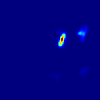} &
        \includegraphics[width=0.95\linewidth,height=0.95\linewidth]{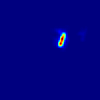} &
        \includegraphics[width=0.95\linewidth,height=0.95\linewidth]{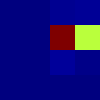} &
        PAN[S]+CTX (attention 4)\\ \cline{2-6}
        &
        \includegraphics[width=0.95\linewidth,height=0.95\linewidth]{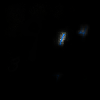} &
        \includegraphics[width=0.95\linewidth,height=0.95\linewidth]{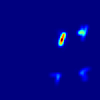} &
        \includegraphics[width=0.95\linewidth,height=0.95\linewidth]{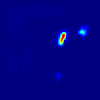} &
        \includegraphics[width=0.95\linewidth,height=0.95\linewidth]{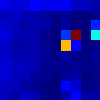} &
        PAN[H]+CTX (attention 3)\\ \cline{2-6}
        &
        \includegraphics[width=0.95\linewidth,height=0.95\linewidth]{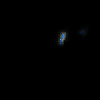} &
        \includegraphics[width=0.95\linewidth,height=0.95\linewidth]{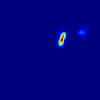} &
        \includegraphics[width=0.95\linewidth,height=0.95\linewidth]{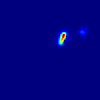} &
        \includegraphics[width=0.95\linewidth,height=0.95\linewidth]{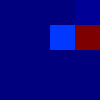} &
        PAN[H]+CTX (attention 4)\\ \cline{2-6}
    \end{tabular}
    \vspace{0.3cm}
    \caption{More qualitative results of SAN, HAN, PAN[S]+CTX and PAN[H]+CTX with query '1'.}
    \label{tab:my_label}
\end{table}

\clearpage
\section{More Qualitative Results on Visual Genome}
\begin{table}[h]
    \centering
    \begin{tabular}{
    @{}C{3cm}@{}|@{}C{3cm}@{}|@{}C{3cm}@{}@{}C{3cm}@{}
    }
        Inputs & SAN & \multicolumn{2}{@{}C{6cm}@{}}{PAN[S]+CTX} \\ \hline
        \includegraphics[width=0.95\linewidth,height=0.95\linewidth]{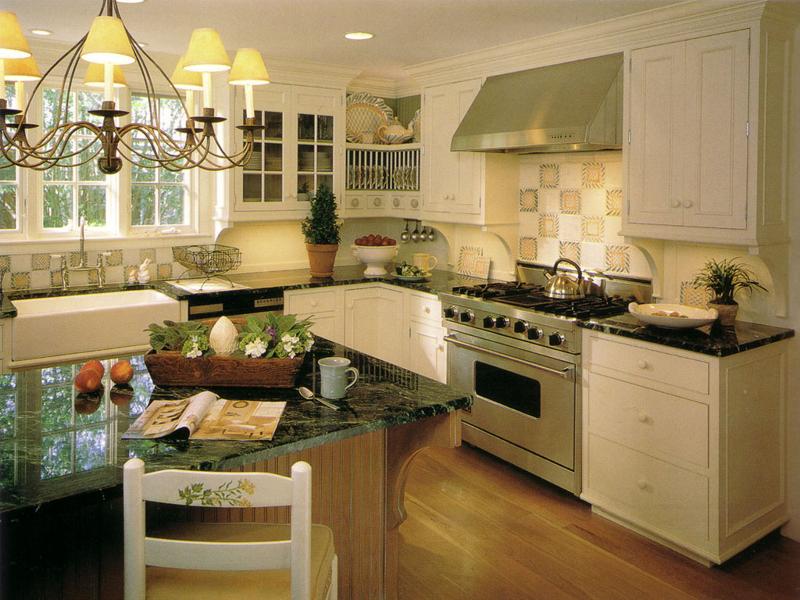} &
        \includegraphics[width=0.95\linewidth,height=0.95\linewidth]{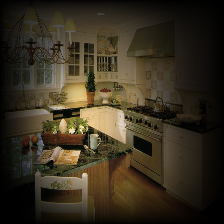} &
        \includegraphics[width=0.95\linewidth,height=0.95\linewidth]{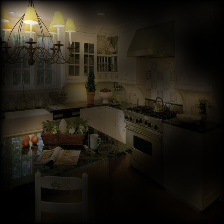} &
        \includegraphics[width=0.95\linewidth,height=0.95\linewidth]{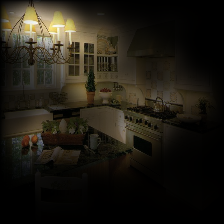} \\
        Query: light &
        \includegraphics[width=0.95\linewidth,height=0.95\linewidth]{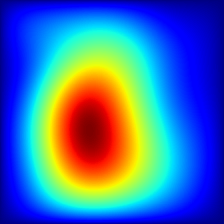} &
        \includegraphics[width=0.95\linewidth,height=0.95\linewidth]{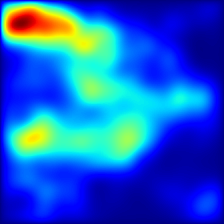} &
        \includegraphics[width=0.95\linewidth,height=0.95\linewidth]{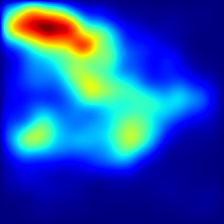} \\ \cline{2-4}
        & HAN & \multicolumn{2}{@{}C{6cm}@{}}{PAN[H]+CTX} \\ \cline{2-4}
        &
        \includegraphics[width=0.95\linewidth,height=0.95\linewidth]{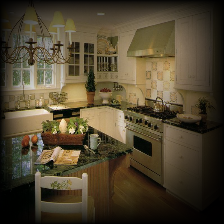} &
        \includegraphics[width=0.95\linewidth,height=0.95\linewidth]{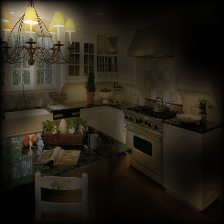} &
        \includegraphics[width=0.95\linewidth,height=0.95\linewidth]{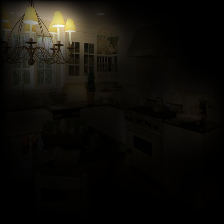} \\
        &
        \includegraphics[width=0.95\linewidth,height=0.95\linewidth]{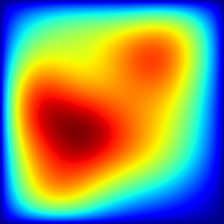} &
        \includegraphics[width=0.95\linewidth,height=0.95\linewidth]{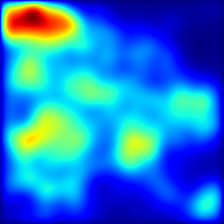} &
        \includegraphics[width=0.95\linewidth,height=0.95\linewidth]{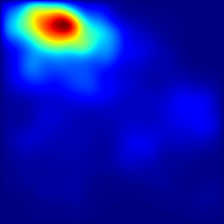} \\
    \end{tabular}
    \vspace{0.3cm}
    \caption{
Attention visualizations of models on VG dataset.
Two variants of progressive attention models gradually attend to target objects in fine resolution. For PAN[$*$]+CTX, we only show last two attentions, which accumulate attentions of earlier layers.
More qualitative results are presented in supplementary document.
}
\end{table}

\begin{table}[h]
    \centering
    \begin{tabular}[m]{
    @{}C{3cm}@{}|@{}C{3cm}@{}|@{}C{3cm}@{}@{}C{3cm}@{}
    }
        Inputs & SAN & \multicolumn{2}{@{}C{6cm}@{}}{PAN[S]+CTX} \\ \hline
        \includegraphics[width=0.95\linewidth,height=0.95\linewidth]{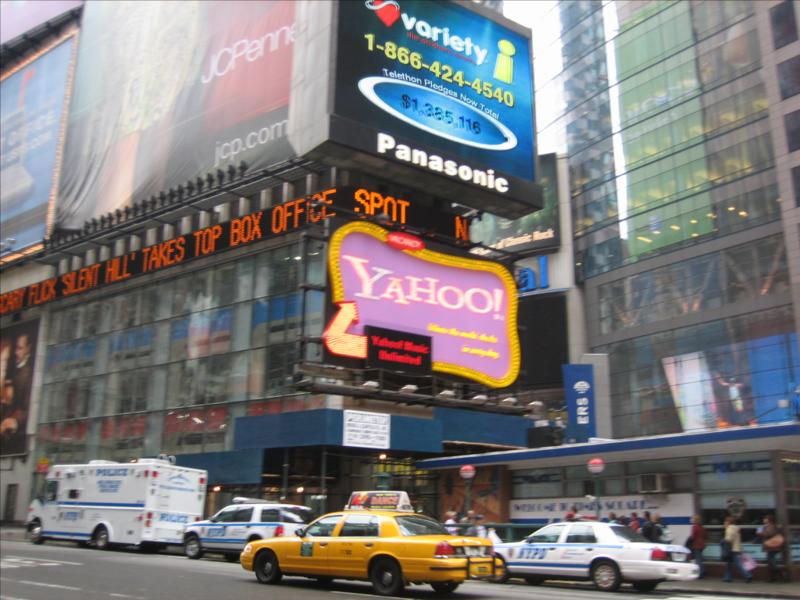} &
        \includegraphics[width=0.95\linewidth,height=0.95\linewidth]{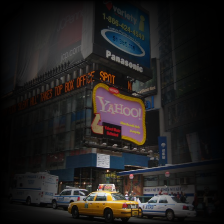} &
        \includegraphics[width=0.95\linewidth,height=0.95\linewidth]{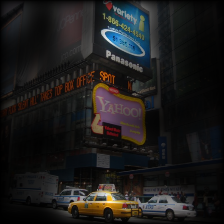} &
        \includegraphics[width=0.95\linewidth,height=0.95\linewidth]{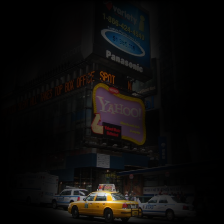} \\
        Query: cap &
        \includegraphics[width=0.95\linewidth,height=0.95\linewidth]{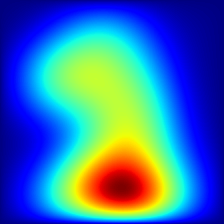} &
        \includegraphics[width=0.95\linewidth,height=0.95\linewidth]{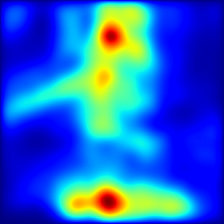} &
        \includegraphics[width=0.95\linewidth,height=0.95\linewidth]{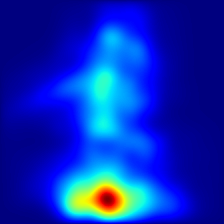} \\ \cline{2-4}
        & HAN & \multicolumn{2}{@{}C{6cm}@{}}{PAN[H]+CTX} \\ \cline{2-4}
        &
        \includegraphics[width=0.95\linewidth,height=0.95\linewidth]{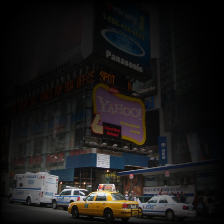} &
        \includegraphics[width=0.95\linewidth,height=0.95\linewidth]{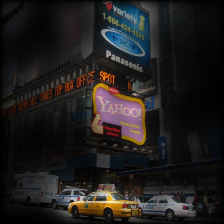} &
        \includegraphics[width=0.95\linewidth,height=0.95\linewidth]{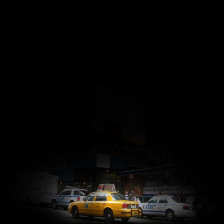} \\
        &
        \includegraphics[width=0.95\linewidth,height=0.95\linewidth]{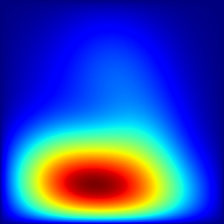} &
        \includegraphics[width=0.95\linewidth,height=0.95\linewidth]{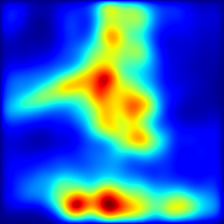} &
        \includegraphics[width=0.95\linewidth,height=0.95\linewidth]{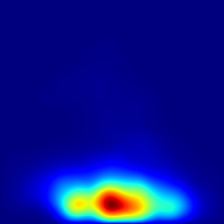} \\
    \end{tabular}
    \vspace{0.3cm}
    \caption{More visualizations of attentions.}
\end{table}

\begin{table}[h]
    \centering
    \begin{tabular}[m]{
    @{}C{3cm}@{}|@{}C{3cm}@{}|@{}C{3cm}@{}@{}C{3cm}@{}
    }
        Inputs & SAN & \multicolumn{2}{@{}C{6cm}@{}}{PAN[S]+CTX} \\ \hline
        \includegraphics[width=0.95\linewidth,height=0.95\linewidth]{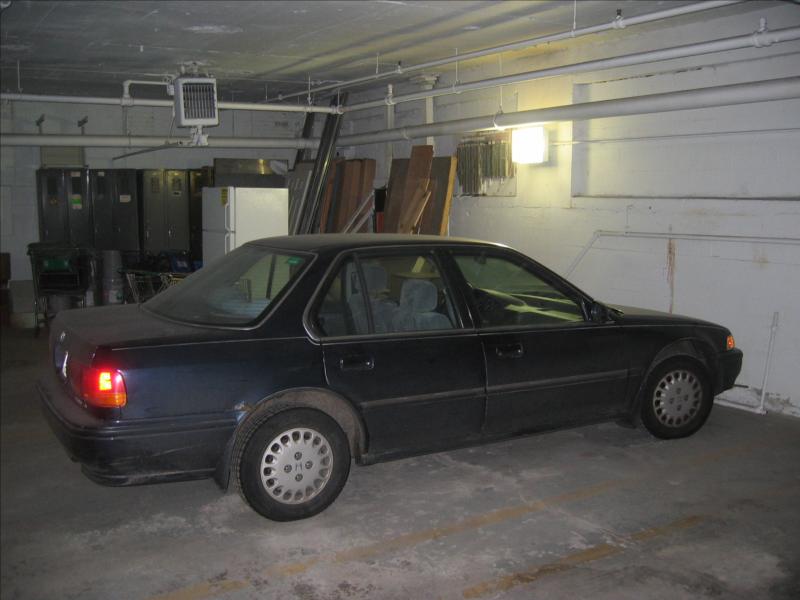} &
        \includegraphics[width=0.95\linewidth,height=0.95\linewidth]{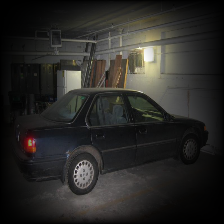} &
        \includegraphics[width=0.95\linewidth,height=0.95\linewidth]{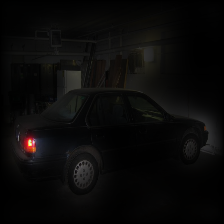} &
        \includegraphics[width=0.95\linewidth,height=0.95\linewidth]{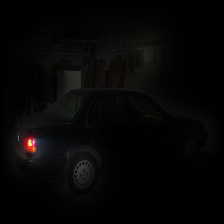} \\
        Query: light &
        \includegraphics[width=0.95\linewidth,height=0.95\linewidth]{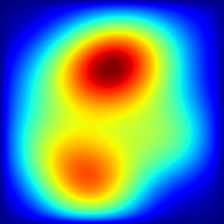} &
        \includegraphics[width=0.95\linewidth,height=0.95\linewidth]{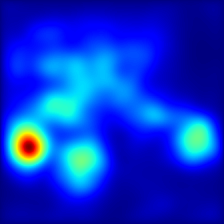} &
        \includegraphics[width=0.95\linewidth,height=0.95\linewidth]{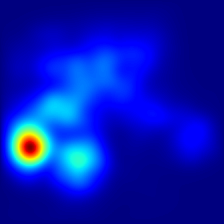} \\ \cline{2-4}
        & HAN & \multicolumn{2}{@{}C{6cm}@{}}{PAN[H]+CTX} \\ \cline{2-4}
        &
        \includegraphics[width=0.95\linewidth,height=0.95\linewidth]{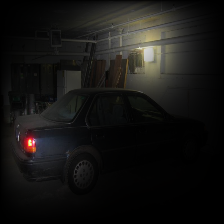} &
        \includegraphics[width=0.95\linewidth,height=0.95\linewidth]{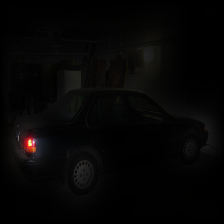} &
        \includegraphics[width=0.95\linewidth,height=0.95\linewidth]{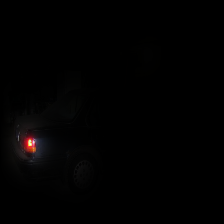} \\
        &
        \includegraphics[width=0.95\linewidth,height=0.95\linewidth]{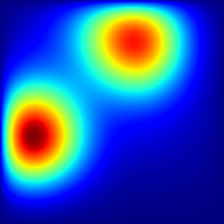} &
        \includegraphics[width=0.95\linewidth,height=0.95\linewidth]{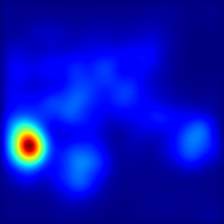} &
        \includegraphics[width=0.95\linewidth,height=0.95\linewidth]{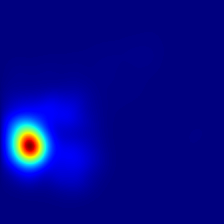} \\
    \end{tabular}
    \vspace{0.3cm}
    \caption{More visualizations of attentions.}
\end{table}
\end{appendices}
\end{document}